\documentstyle[aclap]{article}

\newcommand{\rewrites}{\longrightarrow}

\def\LKP{\mathord{L^2_{K,P}}}
\def\parent{\mathrel{\triangleleft}}

\def\dom{\mathrel{\triangleleft^*}}
\def\eq{\mathrel{\approx}}
\def\lft{\mathrel{\prec}}
\newcommand{\f}[1]{{\mathchoice{\mathop{\mbox{\rm #1}}}%
{\mathop{\mbox{\rm #1}}}{\mathop{\mbox{\scriptsize\rm #1}}}%
{\mathop{\mbox{\scriptsize\rm #1}}}}}
\newcommand{\tup}[1]%
  {\left\langle
#1\right\rangle}
\newenvironment{arblk}%
{\begin{array}[t]{l}}%
{\end{array}}
\let\limp=\rightarrow
\newcommand{\says}[1]{\mbox{---\parbox[t]{0.75\displaywidth}{\bf #1}}}

\newcommand{\xp}[1]{%
\ifmmode\mathchoice{\mathord{\mbox{#1P}}}{\mathord{\mbox{#1P}}}%
{\mathord{\mbox{\scriptsize#1P}}}{\mathord{\mbox{\scriptsize#1P}}}%
\else\mbox{#1P}%
\fi}
\newcommand{\xb}[1]{%
\ifmmode\mathchoice{\overline{\rm #1}}{\overline{\rm #1}}%
{\mbox{\scriptsize$\overline{\rm #1}$}}{\mbox{\scriptsize$\overline{\rm #1}$}}%
\else
\mbox{\rlap{$\overline{\rm #1}$}\phantom{\rm #1}}%
\fi}
\newcommand{\xz}[1]{%
\ifmmode\mathchoice{\mathord{\rm #1}^0}{\mathord{\rm #1}^0}%
{\mathord{\scriptsize\rm #1}^0}{\mathord{\scriptsize\rm #1}^0}%
\else
\mbox{${\rm #1}^0$}%
\fi}
\newcommand{\cat}[2]{%
{
\ifmmode{[_{#1}{#2}]}%
\else$[_{\mbox{\scriptsize #1}}\mbox{#2}]$%
\fi}%
}

\newcommand{\minus}{\ifmmode{-}\else{$-$}\fi}
\newcommand{\Abar}{\mbox{\rlap{$\overline{\rm A}$}\phantom{\rm A}}}
\newcommand{\Rbar}{\mbox{\rlap{$\overline{\rm Ref}$}\phantom{\rm Ref}}}

\def\topmarginnote#1{
}

\title{A Model-Theoretic Framework for Theories of Syntax}

\author{James Rogers\\Institute for Research in Cognitive Science\\
University of Pennsylvania\\Suite 400C, 3401 Walnut Street\\
Philadelphia, PA 19104\\
{\tt jrogers@linc.cis.upenn.edu}
}

\begin{document}

\topmarginnote{In:  Proceedings of the 34th Annual Meeting of the ACL. 
Santa Cruz, USA. 1996.}

\bibliographystyle{fullname}

\maketitle

\begin{abstract}
A natural next step in the evolution of constraint-based grammar
formalisms from rewriting formalisms is to abstract fully away from the details
of the grammar mechanism---to express syntactic theories purely in terms of the
properties of the class of structures they license.  By focusing on the
structural properties of languages rather than on mechanisms for generating or
checking structures that exhibit those properties, this model-theoretic
approach can offer simpler and 
significantly clearer expression of theories and can potentially provide a
uniform formalization, allowing disparate theories to be compared on
the basis of those properties.  We discuss $\LKP$, a monadic second-order
logical framework for such an approach to syntax that has the
distinctive virtue of being superficially expressive---supporting direct
statement of most linguistically significant syntactic properties---but having
well-defined strong generative capacity---languages are definable in $\LKP$ iff
they are strongly context-free.  We draw examples from the realms of GPSG and
GB.
\end{abstract}

\section{Introduction}
Generative grammar and formal language theory share a common origin in
a procedural notion of grammars: the grammar formalism provides
a general mechanism for recognizing or generating languages while the
grammar itself specializes that mechanism for a specific language.  At least
initially there was hope that this relationship would be informative
for linguistics, that by characterizing the natural languages in terms
of language-theoretic complexity one would gain insight into the
structural regularities of those languages.  Moreover, the fact
that language-theoretic complexity classes have dual automata-theoretic
characterizations offered the prospect that such results might provide abstract
models of the human language faculty, thereby  not just identifying these
regularities, but actually accounting for them.

Over time, the two disciplines have
gradually become estranged, principally due to 
a realization that the structural properties of languages
that characterize natural languages may well not be those that can be
distinguished by existing language-theoretic complexity classes.  Thus the
insights offered by formal language theory might actually be misleading in
guiding theories of syntax.  As a
result, the emphasis in generative grammar has turned from formalisms
with restricted generative capacity to those that
support more natural expression of the observed regularities of
languages.  While a variety of distinct approaches have developed, most
of them can be characterized as {\em constraint based}---the formalism
(or formal framework) provides a class of structures and a means of
precisely stating constraints on their form, the linguistic theory is
then expressed as a system of constraints (or principles) that characterize the
class of well-formed analyses of the strings in the language.\footnote{This 
notion of {\em constraint-based\/} includes not only the obvious
formalisms, but the formal framework of GB as well.}

As the study of the formal properties of classes of structures defined in such
a way falls within domain of Model Theory, it's not surprising that
treatments of the meaning of these systems of constraints are typically couched
in terms of formal 
logic~\cite{KR86,MR87,KasRou90,GaEtAl88,johnso88,smolka88,DawVij90,carpen92,%
keller93,RogVS93d}.

While this provides a model-theoretic interpretation of the systems of
constraints produced by these formalisms, those systems are
typically built by derivational processes that employ extra-logical mechanisms
to combine constraints.
More recently, it has become clear that in many cases these mechanisms can
be replaced with ordinary logical operations.  (See, for
instance: \newcite{johnso89}, \newcite{stable92}, \newcite{cornel92},
\newcite{BlGaMe93}, \newcite{BlaMey94}, \newcite{keller93}, \newcite{rogers94},
\newcite{kracht95}, and, anticipating all of these,~\newcite{JohPos80}.) 
This approach abandons
the notions of grammar mechanism and derivation in favor of defining 
languages as classes of more or less ordinary mathematical structures
axiomatized by sets of more or less ordinary logical formulae.  A grammatical
theory expressed within such a framework is just 
the set of logical consequences of those axioms.  This step completes the
detachment of generative grammar from its procedural roots.  Grammars, in this
approach, are purely declarative definitions of a class of structures,
completely independent of mechanisms to generate or check them.
While it is unlikely that every theory of syntax with an explicit derivational
component can be captured in this way,\footnote{Whether there are theories
that cannot be captured, at least without explicitly encoding the derivations,
is an open question of considerable theoretical interest, as is the question of
what empirical consequences such an essential dynamic character might have.}
for those that can the logical re-interpretation frequently offers
a simplified statement of the theory and clarifies its consequences.

But the accompanying loss of language-theoretic complexity results is
unfortunate.  While such results may not be useful in guiding syntactic theory,
they are not irrelevant.  The nature of language-theoretic complexity
hierarchies is to classify languages on the basis of their structural
properties.  The languages in a class, for instance, will typically exhibit
certain closure properties (e.g., pumping lemmas) and the classes themselves
admit normal forms (e.g., representation theorems).  While the linguistic
significance of individual 
results of this sort is open to debate, they at least loosely parallel typical
linguistic concerns:  closure properties state regularities that are exhibited
by the languages in a class, normal forms express generalizations about their
structure.  So while these may not be the right results, they are not entirely
the wrong {\em kind\/} of results.  Moreover, since these classifications are
based on structural properties and the structural properties of natural
language can be studied more or less directly, there is a reasonable
expectation of finding empirical evidence falsifying a hypothesis about
language-theoretic complexity of natural languages if such evidence exists.

Finally, the fact that these complexity classes have automata-theoretic
characterizations means that results concerning the complexity of
natural languages will have implications for the nature of the human language
faculty.  These automata-theoretic characterizations determine, along one axis,
the types of resources required to generate or recognize the languages in a
class.  The regular languages, for instance, can be characterized by
finite-state (string) automata---these languages can be processed
using a fixed amount of memory.  The context-sensitive languages, on the other
had, can be characterized by linear-bounded automata---they can be
processed using 
an amount of memory proportional to the length of the input.  The context-free
languages are probably best characterized by finite-state tree
automata---these 
correspond to recognition by a collection of processes, each with a fixed
amount of memory,  where the number of processes is linear in the length of the
input and all communication between processes is completed at the time they are
spawned.  As a result, while these results do not necessarily offer abstract
models of the human language faculty (since the complexity results do not
claim to characterize the human languages, just to classify them), they do
offer lower bounds on certain abstract properties of that faculty.  In this
way, generative grammar in concert with formal language theory offers
insight into a deep aspect of human 
cognition---syntactic processing---on the basis of observable behavior---the
structural properties of human languages.

In this paper we discuss an approach to defining theories of syntax based on
$\LKP$~\cite{rogers94}, a monadic second-order 
language that has well-defined generative capacity: sets of finite trees are
definable within $\LKP$ iff they are strongly context-free in a particular
sense.  While originally introduced as a means of establishing
language-theoretic complexity results for constraint-based theories, this
language has much to recommend it as a general framework for theories
of syntax in its own right.  Being a monadic second-order language it can
capture the (pure) modal languages of much of the existing model-theoretic
syntax literature directly; 
having a signature based on the traditional linguistic relations of domination,
immediate domination, linear precedence, etc.\ it can express most linguistic
principles transparently; and having a clear characterization in terms of
generative capacity, it serves to re-establish the close connection between
generative grammar and formal language theory that was lost in the move away
from phrase-structure grammars.  Thus,
with this framework we get both the advantages of the model-theoretic
approach with respect to naturalness and clarity in expressing linguistic
principles and the advantages of the grammar-based approach with respect to
language-theoretic complexity results.  

We look, in particular, at the
definitions of a single aspect of each of GPSG and GB.  The first of these,
Feature Specification Defaults in GPSG, are widely assumed
to have an inherently dynamic character.  In addition to being purely
declarative, our reformalization is considerably simplified wrt the definition
in~\newcite{GaKlPuSa85},\footnote{We will refer to~\newcite{GaKlPuSa85} as
GKP\&S} and does not share its 
misleading dynamic flavor.\footnote{We should note that the definition of FSDs
in GKP\&S is, in fact, declarative although this is obscured by the fact that
it is couched in terms of an algorithm for checking models.}  We offer this as
an example of 
how re-interpretations of this sort can inform the original theory.  In the
second example we sketch a definition of chains in GB.  This, again,
captures a 
presumably dynamic aspect of the original theory in a static way.  Here,
though, the main significance of the definition is that it forms a component of
a full-scale treatment of a GB theory of English S- and D-Structure within
$\LKP$.  This full definition establishes that the theory we capture licenses a
strongly context-free language.  More importantly, by examining the limitations
of this definition of 
chains, and in particular the way it fails for examples of non-context-free
constructions, we develop a characterization of the context-free languages that
is quite natural in the realm of GB.  This suggests that the apparent mismatch
between formal language theory and natural languages may well have more to do
with the unnaturalness of the traditional diagnostics than a lack of relevance
of the underlying structural properties.

Finally, while GB and GPSG are fundamentally distinct, even antagonistic,
approaches to syntax, their translation into the model-theoretic terms of
$\LKP$ allows us to explore the similarities between the theories they express
as well as to delineate actual distinctions between them.  We look briefly
at two of these issues.

Together these examples are chosen to illustrate the main strengths of
the model-theoretic approach, at least as embodied in $\LKP$, as a
framework for studying theories of syntax: a focus on structural
properties themselves, rather than on mechanisms for specifying them
or for generating or checking structures that exhibit them, and a
language that is expressive enough to state most linguistically
significant properties in a natural way, but which is restricted
enough to have well-defined strong generative capacity.

\section{\protect{\boldmath{$\LKP$}}---The Monadic Second-Order Language of
Trees} 
$\LKP$ is the monadic second-order language over the signature including a set
of individual constants ($K$), a set of monadic predicates ($P$), and binary
predicates for immediate domination ($\parent$), domination ($\dom$), linear
precedence ($\lft$) and equality ($\eq$).  The predicates in $P$ can be
understood both as picking out particular subsets of the tree and as
(non-exclusive) labels or features decorating the tree.  Models for the
language are labeled {\em tree domains}~\cite{gorn65} with the natural
interpretation of the binary predicates.  In~\newcite{rogers94} we
have shown that 
this language is equivalent in descriptive power to S$\omega$S---the monadic
second-order theory of the complete infinitely branching tree---in the sense
that sets of trees are definable in S$\omega$S iff they are definable in
$\LKP$.  This places it within a hierarchy of results relating
language-theoretic complexity classes to the descriptive complexity of their
models: the sets of strings definable in S1S are exactly the regular
sets~\cite{buchi60}, the sets of finite trees definable in S$n$S, for finite
$n$, are the recognizable sets (roughly the sets of derivation trees of
CFGs)~\cite{doner70}, and, it can be shown, the sets of finite trees definable
in S$\omega$S are those generated by generalized CFGs in which regular
expressions may occur on the rhs of rewrite
rules~\cite{rogers96.dctr}.\footnote{There is reason to believe that this
hierarchy can be extended to encompass, at least, a variety of mildly
context-sensitive languages as well.}  Consequently, languages are definable in
$\LKP$ iff they are strongly context-free in the mildly generalized sense of
GPSG grammars.

In restricting ourselves to the language of $\LKP$ we are restricting ourselves
to reasoning in terms of just the predicates of its signature.  We can expand
this by defining new predicates, even higher-order predicates that express, for
instance, properties of or relations between sets, and in doing so we can use
monadic predicates and individual constants freely since we can interpret these
as existentially bound variables.  But the fundamental restriction of $\LKP$ is
that all predicates other than monadic first-order predicates must be
explicitly defined, that is, their definitions must resolve, via syntactic
substitution, into formulae involving only the signature of $\LKP$.

\section{Feature Specification Defaults in GPSG}
We now turn to our first application---the definition of Feature Specification
Defaults (FSDs) in GPSG.\footnote{A more complete treatment of GPSG in $\LKP$
can be found in \newcite{rogers96.gpsg}.}  Since GPSG is presumed to license
(roughly) context-free languages, we are not concerned here with establishing
language-theoretic complexity but rather with clarifying the linguistic theory
expressed by GPSG.  FSDs specify conditions on feature values that must hold at
a node in a licensed tree unless they are overridden by some other component of
the grammar; in particular, unless they are incompatible with either a feature
specified by the ID rule licensing the node ({\em inherited\/} features) or a
feature required by one of the agreement principles---the Foot Feature
Principle (FFP), Head Feature Convention (HFC), or Control Agreement Principle
(CAP).  It is the fact that the default holds just in case it is incompatible
with these other components that gives FSDs their dynamic flavor.  Note,
though, in contrast to typical applications of default logics, a GPSG grammar
is not an evolving theory.  The exceptions to the defaults are fully determined
when the grammar is written.  If
we ignore for the moment the effect of the agreement principles, the defaults
are roughly the converse of the ID rules: a non-default feature occurs iff it
is licensed by an ID rule.

It is easy to capture ID rules in $\LKP$.  For instance the rule:
\[
\xp{V} \rewrites \f{H}[5],\,\xp{N},\,\xp{N}
\]
can be expressed:
\[
\begin{array}{l}
\f{ID}_5(x,y_1,y_2,y_3) \equiv\\
\quad\f{Children}(x,y_1,y_2,y_3)\land\xp{V}(x)\land\\
\quad\f{H}(y_1)\land\tup{\f{SUBCAT},5}(y_1)\land\xp{N}(y_2)\land
\xp{N}(y_3),
\end{array}
\]
where $\f{Children}(x,y_1,y_2,y_3)$ holds iff the set of nodes that are
children of $x$ are just the $y_i$ and \xp{V}, $\tup{\f{SUBCAT},5}$, etc.\ are
all members of $P$.\footnote{We will not elaborate here on the encoding of
categories in $\LKP$, nor on non-finite ID schema like the iterating
co-ordination schema.  These present no significant problems.}  A sequence of
nodes will satisfy $\f{ID}_5$ iff they form a local tree that, in the
terminology of GKP\&S, is {\em induced\/} by the corresponding ID rule.  Using
such encodings we can define a predicate $\f{Free}_f(x)$ which is true at a
node $x$ iff the feature $f$ is compatible with the inherited features of $x$.

The agreement principles require pairs of nodes occurring in certain
configurations in local trees to agree on certain classes of features.  Thus
these principles do not introduce features into the trees, but rather propagate
features from one node to another, possibly in many steps.  Consequently, these
principles cannot override FSDs by themselves; rather every violation of a
default must be licensed by an inherited feature somewhere in the tree.  In
order to account for this propagation of features, the definition of FSDs in
GKP\&S is based on 
identifying pairs of nodes that co-vary wrt the relevant features in all
possible extensions of the given tree.  As a result, although the treatment in
GKP\&S is actually declarative, this fact is far from obvious.

Again, it is not difficult to define the configurations of local trees in which
nodes are required to agree by FFP, CAP, or HFC in $\LKP$.  Let the predicate
$\f{Propagate}_f(x,y)$ hold for a pair of nodes $x$ and $y$ iff they are
required to agree on $f$ by one of these principles (and are, thus, in the
same local tree).  Note that $\f{Propagate}$ is symmetric.  Following the
terminology of GKP\&S, we can identify the set of nodes that are prohibited 
from taking feature $f$ by the combination of the ID rules, FFP, CAP, and HFC
as the set of nodes that are {\em privileged\/} wrt $f$.  This includes all
nodes that are not Free for $f$ as well as any node connected to such a node by
a sequence of $\f{Propagate}_f$ links.  We, in essence, define this
inductively.  $\f{P}'_f(X)$ is true of a set iff it includes all nodes not Free
for $f$ and is closed wrt $\f{Propagate}_f$.  $\f{PrivSet}_f(X)$ is true of
the smallest such set.
\[
\begin{array}{rcl}
\multicolumn{3}{l}{\f{P}'_f(X) \equiv}\\ 
&&(\forall x)[\neg\f{Free}_f(x)\limp X(x)]\;\land\\
& & (\forall x)[(\exists y)[X(y)\land\f{Propagate}_f(x,y)]\limp X(x)]\\
\end{array}
\]
\[
\begin{array}{rcl}
\f{PrivSet}_f(X) &\equiv&
\f{P}'_f(X)\;\land\\
&&(\forall Y)[\f{P}'_f(Y)\limp\f{Subset}(X,Y)].
\end{array}
\]
There are two things to note about this definition.  First, in any tree there
is a unique set satisfying $\f{PrivSet}_f(X)$ and this contains exactly those
nodes not Free for $f$ or connected to such a node by $\f{Propagate}_f$.
Second, while this is a first-order inductive property, the definition is a
second-order explicit definition.  In fact, the second-order quantification of
$\LKP$ allows us to capture any monadic first-order inductively or implicitly
definable property explicitly.

Armed with this definition, we can identify individuals that are privileged wrt
$f$ simply as the members of $\f{PrivSet}_f$.\footnote{We could, of course,
skip the definition of $\f{PrivSet}_f$ and define $\f{Privileged}_f(x)$ as
$(\forall X)[\f{P}'(X)\limp X(x)]$, but we prefer to emphasize the inductive
nature of the definition.}

\[
\f{Privileged}_f(x) \equiv (\exists X)[\f{PrivSet}_f(X)\land X(x)].
\]

One can define $\f{Privileged}_{\neg f}(x)$ which holds whenever $x$ is
required to take the feature $f$ along similar lines.

These, then, let us capture FSDs.  For the default 
\cat{}{\minus INV}, for instance, we get:
\[ (\forall x)[\neg\f{Privileged}_{\cat{}{\minus\,\f{INV}}}(x)\limp
\cat{}{\minus\,\f{INV}}(x)].
\]
For $\cat{}{\f{BAR}\,0}\supset\sim\cat{}{\f{PAS}}$ (which says that 
\cat{}{Bar 0} nodes are, by default, not marked passive), we get:
\[
(\forall x)[\begin{arblk}
(\cat{}{\f{BAR}\,0}(x)\land
  \neg\f{Privileged}_{\neg\cat{}{\f{PAS}}}(x))\limp\\
\neg\cat{}{\f{PAS}}(x)].\end{arblk}
\]

The key thing to note about this treatment of FSDs is its simplicity relative
to the treatment of GKP\&S.  The second-order quantification allows us to
reason directly in terms of the sequence of nodes extending from the privileged
node to the local tree that actually licenses the privilege.  The immediate
benefit is the fact that it is clear that the property of satisfying a set of
FSDs is a static property of labeled trees and does not depend on the
particular strategy employed in checking the tree for compliance.

\section{Chains in GB}
The key issue in capturing GB theories within $\LKP$ is the fact that the
mechanism of free-indexation is provably non-definable.  Thus definitions of
principles that necessarily employ free-indexation have no direct
interpretation in $\LKP$ (hardly surprising, as we expect GB to be capable of
expressing non-context-free languages).  In many cases, though, references to
indices can be eliminated in favor of the underlying structural relationships
they express.\footnote{More detailed
expositions of the interpretation of GB in $\LKP$ can be found in
\newcite{rogers96}, \newcite{rogers95}, and \newcite{rogers94}.}
The most prominent example is the definition of the {\em chains\/}
formed by move-$\alpha$.  The fundamental problem here is identifying each
trace with its antecedent without referencing their index.  Accounts of the
licensing of traces that, in many cases of movement, replace co-indexation with
government relations have been offered by both~\newcite{rizzi90}
and~\newcite{manzini92}.  The key element of these accounts, from our point 
of view, is that the antecedent of a trace must be the closest
antecedent-governor of the appropriate type.  These relationships are easy to
capture in $\LKP$.  For \Abar-movement, for instance, we have:
\begin{eqnarray*}
\lefteqn{\f{\Abar-Antecedent-Governs}(x,y) \equiv}
\\
&&
\neg\f{A-pos}(x)\land\f{C-Commands}(x,y)\land\f{F.Eq}(x,y)\land
\\
&&\says{$x$ is a potential antecedent in an \Abar-position}\\
&&
\neg(\exists z)[\f{Intervening-Barrier}(z,x,y)]\land\\
&&\says{no barrier intervenes}\\
&&
\neg(\exists z)[\f{Spec}(z)\land\neg\f{A-pos}(z)\land\\
&&
\hphantom{\neg(\exists z)[}\f{C-Commands}(z,x)\land
\f{Intervenes}(z,x,y)]\\
&&\says{minimality is respected}
\end{eqnarray*}
where $\f{F.Eq}(x,y)$ is a conjunction of biconditionals that assures that $x$
and $y$ agree on the appropriate features and the other predicates are are
standard GB notions that are definable in $\LKP$.

Antecedent-government, in Rizzi's and Manzini's accounts, is the key
relationship between adjacent members of chains which are identified by
non-referential indices, but plays no role in the definition of chains which
are assigned a referential index.\footnote{This accounts for subject/object
asymmetries.}  Manzini argues, however, that referential chains cannot overlap,
and thus we will never need to distinguish multiple referential chains in any
single context.  Since we can interpret any bounded number of indices simply as
distinct labels, there is no difficulty in identifying the members of
referential chains in $\LKP$.  On these and similar grounds we can extend these
accounts to identify adjacent members of referential chains, and, at least in
the case of English, of chains of head movement and of rightward movement.
This gives us five mutually exclusive relations which we can combine into a
single {\em link\/} relation that must hold between every trace and its
antecedent:
\begin{eqnarray*}
\f{Link}(x,y) &\equiv&\f{A-Link}(x,y)\lor\f{\Abar-\Rbar-Link}(x,y)\lor
\\
&&\f{\Abar-Ref-Link}(x,y)\lor\f{\xz{X}-Link}(x,y)\lor\\
&&\f{Right-Link}(x,y).
\end{eqnarray*}

The idea now is to define chains as sequences of nodes that are linearly
ordered by $\f{Link}$, but before we can do this there is still one issue to
resolve.  While minimality ensures that every trace must have a unique
antecedent, we may yet admit a single antecedent that licenses multiple traces.
To rule out this possibility, we require chains to be {\em closed\/} wrt the
link relation, i.e., every chain must include every node that is related by
$\f{Link}$ to any node already in the chain.  Our definition, then, is in
essence the definition, in GB terms, of a discrete linear order with endpoints,
augmented with this closure property.
\begin{eqnarray*}
\lefteqn{\f{Chain}(X) \equiv}\\
&&(\exists! x)[X(x)\land\f{Target}(x)]\land\\
&&\says{$X$ contains exactly one Target}\\
&&(\exists! x)[X(x)\land\f{Base}(x)]\land\\
&&\says{and one Base}\\
&&
(\forall x)[X(x)\land\neg\f{Target}(x)\limp\\
&&\hphantom{(\forall x)[} (\exists!y)[X(y)\land\f{Link}(y,x)]]\quad\land\\
&&\says{All non-Target have a unique antecedent in $X$}\\
&&
(\forall x)[X(x)\land\neg\f{Base}(x)\limp\\
&&\hphantom{(\forall x)[}  (\exists!y)[X(y)\land\f{Link}(x,y)]]\quad\land\\
&&\says{All non-Base have a unique successor in $X$}\\
&&
(\forall x,y)[X(x)\land(\f{Link}(x,y)\lor\f{Link}(y,x))\limp\\
&&\hphantom{(\forall x,y)[} X(y)]\\
&&\says{$X$ is closed wrt the Link relation}
\end{eqnarray*}
Note that every node will be a member of exactly one (possibly trivial) chain.

The requirement that chains be closed wrt $\f{Link}$ means that
chains cannot overlap unless they are of distinct types.  This definition works
for English because it is possible, in English, to resolve chains
into boundedly many types in such a way that no two chains of the same type
ever overlap.  In fact, it fails only in cases, like head-raising in Dutch,
where there are potentially unboundedly many chains that may overlap a single
point in the tree.  Thus, this gives us a property separating GB
theories of movement that license strongly context-free languages from
those that potentially don't---if we can establish a fixed bound on
the number of chains that can overlap, then the definition we sketch
here will suffice to capture the theory in $\LKP$ and, consequently,
the theory licenses only strongly context-free languages.  This is a reasonably
natural diagnostic for context-freeness in GB and is close to common
intuitions of what is difficult about head-raising constructions; it gives
those intuitions theoretical substance and provides a reasonably clear strategy
for establishing context-freeness. 

\section{A Comparison and a Contrast}
Having interpretations both of GPSG and of a GB account of English
in $\LKP$ provides a certain amount of insight into the distinctions
between these approaches.  For example, while the explanations of
filler-gap relationships in GB and GPSG are quite
dramatically dissimilar, when one focuses on the structures these
accounts license one finds some surprising parallels.  In the light of
our interpretation of antecedent-government, one can understand the
role of minimality in Rizzi's and Manzini's accounts as eliminating
ambiguity from the sequence of relations connecting the gap with its
filler.  In GPSG this connection is made by the sequence of agreement
relationships dictated by the Foot Feature Principle.  So while both
theories accomplish agreement between filler and gap through marking a
sequence of elements falling between them, the GB account marks
as few as possible while the GPSG account marks every node of the
spine of the tree spanning them.  In both cases, the complexity of the
set of licensed structures can be limited to be strongly context-free
iff the number of relationships that must be distinguished in a given
context can be bounded.

One finds a strong contrast, on the other hand, in the way in which GB
and GPSG encode language universals.  In GB it is presumed that
all principles are universal with the theory being specialized to
specific languages by a small set of finitely varying parameters.
These principles are simply properties of trees.  In terms of models,
one can understand GB to define a universal language---the set of all
analyses that can occur in human languages.  The principles then
distinguish particular sub-languages---the head-final or the {\em pro}-drop
languages, for instance.  Each realized human language is just the
intersection of the languages selected by the settings of its
parameters.  In GPSG, in contrast, many universals are, in
essence, closure properties that must be exhibited by human
languages---if the language includes trees in which a particular
configuration occurs then it includes variants of those trees in which
certain related configurations occur.  Both the ECPO principle and
the metarules can be understood in this way.  Thus while
universals in GB are properties of trees, in GPSG they tend to be
properties of {\em sets\/} of trees.  This makes a significant
difference in capturing these theories model-theoretically; in the GB
case one is defining sets of models, in the GPSG case one is defining
sets of sets of models.  It is not at all clear what the linguistic
significance of this distinction is; one particularly interesting question is
whether it has empirical consequences.  It is only from the
model-theoretic perspective that the question even arises.

\section{Conclusion}
We have illustrated a general formal framework for expressing theories of
syntax based on axiomatizing classes of models in $\LKP$.  This approach has a
number of strengths.  First, as should be clear from our brief explorations of
aspects of GPSG and GB, re-formalizations of existing theories 
within $\LKP$ can offer a clarifying perspective on those theories, and, in
particular, on the consequences of individual components of those theories.
Secondly, the framework is purely declarative and focuses on those aspects of
language that are more or less directly observable---their structural
properties.  It allows us to reason about the consequences of a theory without
hypothesizing a specific mechanism implementing it.  The abstract properties of
the mechanisms that might implement those theories, however, are not beyond our
reach.  The key virtue of descriptive complexity results like the
characterizations of language-theoretic complexity classes discussed here and
the more typical characterizations of computational complexity
classes~\cite{gurevi88,immerm89} is that 
they allow us to determine the complexity of checking properties independently
of how that checking is implemented.  Thus we can use such descriptive
complexity results to draw conclusions about those
abstract properties of such mechanisms that are actually inferable from their
observable behavior.  Finally, by providing a uniform representation for a
variety of linguistic theories, it offers a framework for comparing their
consequences.  Ultimately it has the potential to reduce distinctions between
the mechanisms underlying those theories to distinctions between the properties
of the sets of structures they license.  In this way one might hope to
illuminate the empirical consequences of these distinctions, should any, in
fact, exist.


\end{document}